# Benchmarking Vision Transformers and CNNs for Thermal Photovoltaic Fault Detection with Explainable AI Validation

Serra Aksoy

Institute of Computer Science, Ludwig Maximilian University of Munich (LMU), Oettingenstrasse 67, 80538 Munich, Germany;

serra.aksoy@campus.lmu.de, serurays@gmail.com

**Abstract**
Artificial intelligence deployment for automated photovoltaic (PV) monitoring faces interpretability barriers that limit adoption in energy infrastructure applications. While deep learning achieves high accuracy in thermal fault detection, validation that model decisions align with thermal physics principles remains lacking, creating deployment hesitancy where understanding model reasoning is critical. This study provides a systematic comparison of convolutional neural networks (ResNet-18, EfficientNet-B0) and vision transformers (ViT-Tiny, Swin-Tiny) for thermal PV fault detection, using XRAI saliency analysis to assess alignment with thermal physics principles. This represents the first systematic comparison of CNNs and vision transformers for thermal PV fault detection with physics-validated interpretability. Evaluation on 20,000 infrared images spanning normal operation and 11 fault categories shows that Swin Transformer achieves the highest performance (94% binary accuracy; 73% multiclass accuracy) compared to CNN approaches. XRAI analysis reveals that models learn physically meaningful features, such as localized hotspots for cell defects, linear thermal paths for diode failures, and thermal boundaries for vegetation shading, consistent with expected thermal signatures. However, performance varies significantly across fault types: electrical faults achieve strong detection (F1-scores >0.90) while environmental factors like soiling remain challenging (F1-scores 0.20-0.33), indicating limitations imposed by thermal imaging resolution. The thermal physics-guided interpretability approach provides methodology for validating AI decision-making in energy monitoring applications, addressing deployment barriers in renewable energy infrastructure.

**Keywords:** Solar panel fault detection, thermal infrared imaging, deep learning, vision transformers, swin transformer, convolutional neural networks, explainable artificial intelligence, photovoltaic systems, automated inspection, XRAI saliency analysis, renewable energy diagnostics

## 1. Introduction

The global transition toward renewable energy infrastructure has positioned photovoltaic systems as critical components of modern energy security, with solar installations requiring robust monitoring frameworks to ensure reliable operation at scale. As these systems become integral to energy infrastructure, their classification as critical infrastructure warrants stringent evaluation procedures, enhanced protection measures, and regulatory oversight, particularly given their strategic role in supporting energy diversification, environmental sustainability, and economic development. However, the deployment of artificial intelligence for automated monitoring faces fundamental barriers related to interpretability and trustworthiness in safety-critical applications, where stakeholder confidence depends on understanding model decision-making processes rather than relying solely on accuracy metrics. Current deep learning approaches for infrastructure fault detection, while achieving high accuracy, lack validation mechanisms that align model decisions with established physical principles, creating deployment hesitancy in critical systems where failures can have significant consequences[1].

The economic implications are substantial, as photovoltaic installations represent critical infrastructure investments where equipment failures directly impact energy yield and revenue generation. Analysis of operational data from 15 utility-scale installations demonstrates that failure energy losses (FEL) can reach 0.96% of net energy yield, with electrical grid and transformer station failures accounting for 68% of all failure-related energy losses despite representing only 6.48% of total failures[2]. Moreover, these studies reveal that solar field failures represent only 4.26% of all failure energy losses, highlighting that early detection of high-impact electrical faults provides disproportionate economic benefits compared to module-level monitoring alone.

Existing automated inspection methods have demonstrated substantial technical capabilities across various deep learning architectures, yet persistent gaps remain between high performance and practical deployment. Masita et al. (2025) systematically reviewed CNN applications in PV diagnostics, revealing that various models including MobileNet, VGG-16, and YOLO achieve remarkable fault detection performance, yet identified persistent challenges regarding model generalization across different environmental conditions. Boubaker et al. (2023) demonstrated through comprehensive evaluation that deep learning techniques consistently outperform traditional machine learning methods, achieving 98.71% accuracy in fault detection using infrared thermography, while Bu et al. (2023) developed CNN architectures that achieved 97.42% fault classification accuracy[3–5]. Recent specialized architectures have further emphasized this pattern: Dhimish (2025) introduced HOTSPOT-YOLO, achieving 90.8% mean average precision, Bommes et al. (2021) achieved 73.3% to 96.6% AUROC scores through supervised contrastive learning, and Vlaminck et al. (2022) developed region-based CNN approaches achieving over 90% true positive rates. Despite these impressive accuracy metrics across multiple studies, a critical interpretability gap persists that limits industrial adoption in safety-critical applications[6–8].

Trustworthy AI frameworks have identified the essential requirements for deploying AI systems in critical infrastructure, yet existing approaches lack the domain-specific validation mechanisms necessary for energy systems. Li et al. (2021) established comprehensive principles for trustworthy AI deployment, identifying that beyond predictive accuracy, AI systems must demonstrate robustness, explainability, transparency, reproducibility, generalization, fairness, privacy protection, and accountability throughout their operational lifecycle[9]. Their systematic framework emphasizes that "the limitation of an accuracy-only measurement has been exposed to a number of new challenges" and that trustworthy AI requires validation across multiple dimensions simultaneously rather than optimizing individual metrics in isolation. Wang et al. (2023) emphasized that trustworthy edge intelligence systems must possess endogenous security, reliability, transparency, and sustainability characteristics, particularly highlighting that "complex black-box models may entail sacrificing real-time performance" when interpretability requirements are balanced against operational constraints[10]. The increasing connectivity and reliance on digital technologies in industrial and utility-scale PV systems exposes them to sophisticated cyber threats, including unauthorized remote access, malware, ransomware, and denial-of-service attacks that can overwhelm smart inverters and compromise monitoring capabilities. Machlev et al. (2022) highlighted that despite widespread adoption and outstanding performance, machine learning models are considered as "black boxes", since it is very difficult to understand how such models operate in practice, and in the power systems field, which requires a high level of accountability, it is hard for experts to trust and justify decisions and recommendations made by these models[11]. The authors emphasized that power systems planning, and operation is done solely by power experts, based on their knowledge in power systems, supporting programs, and field experience, which is gathered over time, making experts in the power system field find it hard to trust the decisions and recommendations made by machine learning based algorithms, limiting their practical use. This challenge is particularly acute because current techniques provide explanations that are designed for AI experts instead of power systems experts, with the most advanced XAI algorithms developed by computer scientists and AI researchers providing explanations through heat-maps that provide good intuition but do not always contain enough information for the user. Furthermore, a critical challenge exists in how to prevent XAI methods from outputting misleading explanations and how to provide trustworthy recommendations, as explanations can increase the user's trust, but in the long term, the outputs of the models may not make accurate recommendations, potentially causing users to develop incorrect confidence and trust mistaken results. Wang et al. (2023) further identified that interpretable techniques must "strike a balance, ensuring that explanatory information is both comprehensive and adheres to computational constraints" in resource-limited edge environments, while addressing the fundamental challenge that "increasing interpretability often introduce additional computational and storage costs." Li et al. (2021) specifically emphasized that explainability and transparency represent distinct but interconnected requirements, where explainability addresses understanding how AI models make decisions while transparency considers AI as a software system requiring disclosure of information regarding its entire lifecycle, noting that "the opaqueness of complex AI systems has led to widespread concerns in academia, the industry, and society at large." The fundamental limitation is that while existing XAI frameworks identify general principles for explainable AI, they lack mechanisms to validate whether model decisions align with established physical principles governing fault manifestations in specific domains[9,10,12].

Despite extensive research in explainable AI approaches, current methods remain insufficient for critical infrastructure applications where understanding model reasoning is as important as achieving high performance. Ledmaoui et al. (2024) addressed this challenge by developing an explainable AI model that achieved 91.46% accuracy while implementing user-friendly interfaces for decision-making support, demonstrating the importance of interpretability in practical deployments[13]. However, existing explainable AI approaches lack mechanisms to validate model decisions against known physical signatures of infrastructure faults. Chung et al. (2024) further highlighted this challenge in their comprehensive analysis of XAI limitations, demonstrating that current explainability methods often provide misleading explanations and create a "false sense of security" rather than genuine understanding of model behavior[14]. Their work revealed that many state-of-the-art XAI techniques fail under adversarial conditions and suffer from inconsistency issues that undermine their reliability in critical applications. Ali et al. (2023) reinforced these concerns through their extensive survey of explainable AI methods, identifying fundamental gaps between general XAI principles and domain-specific requirements, particularly emphasizing that effective explainability must be tailored to specific application domains and user needs rather than relying on generic interpretation techniques[15]. These challenges are compounded by the critical infrastructure nature of modern photovoltaic systems, which require enhanced security frameworks alongside interpretable AI solutions[16].

The deployment barrier centers on the fundamental question of whether AI systems correctly identify faults based on physically meaningful thermal signatures or rely on spurious correlations that may fail under varying operational conditions. While current approaches excel at pattern recognition, they provide limited assurance that learned features correspond to actual thermal physics principles governing fault manifestations. This limitation becomes critical in infrastructure monitoring where incorrect classifications can lead to unnecessary maintenance interventions or, more dangerously, missed critical failures. Moreover, the recent emergence of vision transformer architectures offers potential advantages over traditional convolutional approaches, yet systematic comparisons for thermal fault detection in energy infrastructure remain unexplored.

This study addresses these deployment barriers through a comprehensive evaluation that combines architectural benchmarking with physics-guided interpretability validation. To our knowledge, we present the first unified, large-scale benchmark comparing convolutional neural networks (ResNet-18, EfficientNet-B0) and vision transformers (ViT-Tiny, Swin-Tiny) for thermal PV fault detection on the Raptor-Maps ~20k IR dataset, evaluated across both binary anomaly detection and multiclass fault classification. This represents the first systematic comparison of CNNs and vision transformers for thermal PV fault detection with physics-validated interpretability. We further implement XRAI saliency to test whether model attention aligns with expected thermal-physics signatures, validating that high-performing models learn physically meaningful features rather than spurious cues. We then analyze architectural trade-offs and performance limitations that inform deployment decisions, and we quantify limits imposed by thermal imaging quality, highlighting fault categories where current approaches excel versus those likely requiring alternative sensing modalities.

## 2. Methodology

### 2.1 Data Acquisition and Preprocessing

This research worked with a large thermal infrared image dataset of solar photovoltaic modules, consisting of 20,000 high-resolution thermal images taken by infrared thermography[17]. The database presents a large range of both normal operating conditions and a multitude of fault modes typically found in actual solar installations, classified into 12 different classes: one normal condition (No-Anomaly) and 11 fault classes including Cell defects, Hot-Spot occurrence, Offline-Module conditions, Vegetation shading, Diode failures, Shadowing effect,

Cracking damage, Soiling deposition, and multi-occurrence variants (Diode-Multi, Hot-Spot-Multi, Cell-Multi). Figure 1 shows representative thermal signatures of each class, which indicate the unique temperature profiles that characterize various fault conditions.

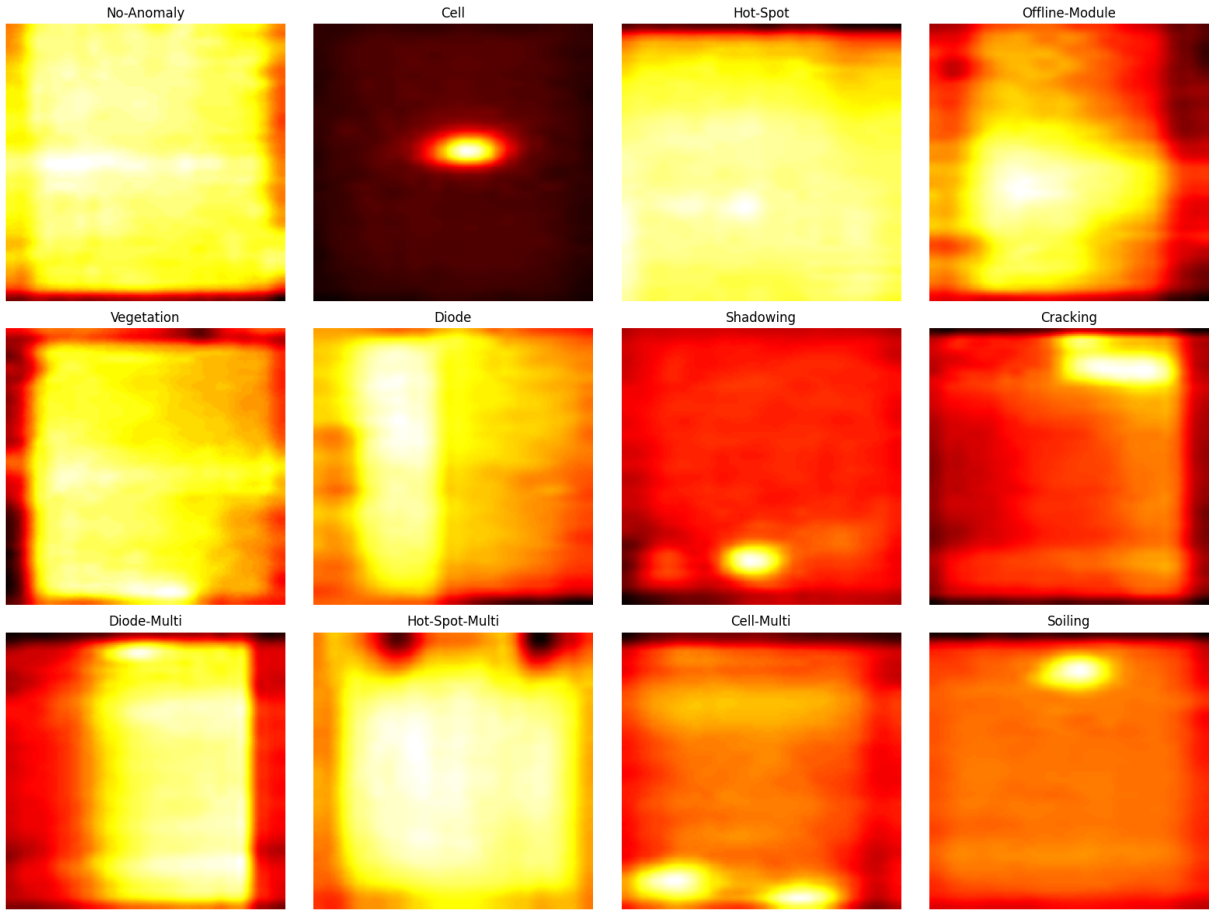

**Figure 1.** Class-wise Visualization of Infrared Module Types

All the thermal images were subjected to standardization procedures to have uniform input dimensions and maintain data integrity. The images were transformed into grayscale mode and resized to 128×128 pixels through cubic interpolation to retain thermal gradient details. The pixel values were normalized between [0,1] through division by 255 to facilitate stable gradient flow in the process of training the neural network. A systematic label encoding scheme was employed to transform categorical fault labels into numerical representations with each fault class being allocated a distinct integer identifier.

Two complementary classification schemes were designed to meet different industrial requirements. The binary classification scheme categorizes solar panel states into normal operation (No-Anomaly, class 0) and anomalous operation (any fault mode, class 1), yielding a perfectly balanced dataset with exactly 10,000 samples per class for anomaly detection tasks. The multiclass classification model is aimed at the fault discrimination alone by removing all the normal operation samples and categorizing the 11 different fault types through zero-indexed classification (labels 0-10), enabling precise fault type determination to aid maintenance decision-making.

Both models employ stratified random sampling with a 70:15:15 training/validation/testing split with proportional class distribution in each split. The binary classification provides 14,000 training, 3,000 validation, and 3,000 test examples with perfect class balance maintained at all times (Table 1).

**Table 1.** Binary Classification Dataset Distribution

| Split | No-Anomaly | Anomaly | Total |
|---|---|---|---|
| **Training** | 7,000 | 7,000 | 14,000 |
| **Validation** | 1,500 | 1,500 | 3,000 |
| **Test** | 1,500 | 1,500 | 3,000 |
| **Total** | 10,000 | 10,000 | 20,000 |

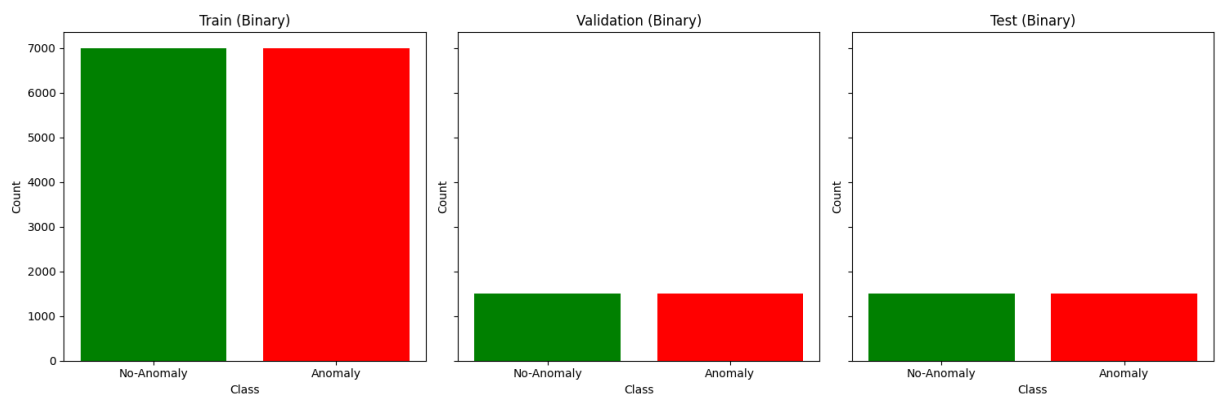

**Figure 2.** Binary Class Distribution Across Train, Validation, and Test Sets

The multiclass classification splits the fault-only dataset of 10,000 samples into 7,000 training, 1,500 validation, and 1,500 test samples with the natural class distribution reflecting actual fault occurrence patterns replicated (Table 2).

**Table 2:** Multiclass Classification Dataset Distribution

| Fault Class | Training | Validation | Test | Total | Percentage |
|---|---|---|---|---|---|
| **Cell** | 1,314 | 282 | 281 | 1,877 | 18.8% |
| **Vegetation** | 1,147 | 246 | 246 | 1,639 | 16.4% |
| **Diode** | 1,049 | 225 | 225 | 1,499 | 15.0% |
| **Cell-Multi** | 902 | 193 | 193 | 1,288 | 12.9% |
| **Shadowing** | 739 | 158 | 159 | 1,056 | 10.6% |
| **Cracking** | 658 | 141 | 141 | 940 | 9.4% |
| **Offline-Module** | 579 | 124 | 124 | 827 | 8.3% |
| **Hot-Spot** | 174 | 38 | 37 | 249 | 2.5% |
| **Hot-Spot-Multi** | 172 | 37 | 37 | 246 | 2.5% |
| **Soiling** | 143 | 30 | 31 | 204 | 2.0% |
| **Diode-Multi** | 123 | 26 | 26 | 175 | 1.8% |
| **Total** | 7,000 | 1,500 | 1,500 | 10,000 | 100.0% |

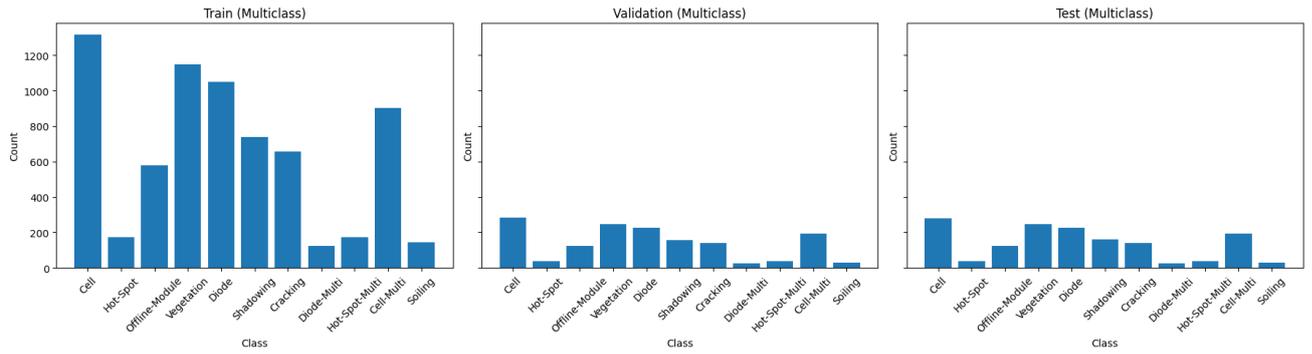

**Figure 3.** Multiclass Distribution Across Train, Validation, and Test Splits

Figures 2 and 3 illustrate the class distribution of the training, validation, and test sets for binary and multiclass classifications, respectively, reflecting the balanced nature of the binary classification system and the complex issue of class imbalance in the multiclass classification system.

### 2.2 Experimental Setup

Four of the contemporary deep learning architectures were contrasted for both binary and multiclass classification tasks: ResNet-18, EfficientNet-B0, Vision Transformer (ViT-Tiny), and Swin Transformer (Swin-Tiny). These architectures embody various computational paradigms, such as conventional residual learning, compound scaling principles, pure attention mechanisms, and hierarchical attention architectures. The experiment procedure consisted of binary anomaly detection for distinguishing between normal and defective panels and multiclass fault identification related to recognizing eleven categories of faults (Figure 4).

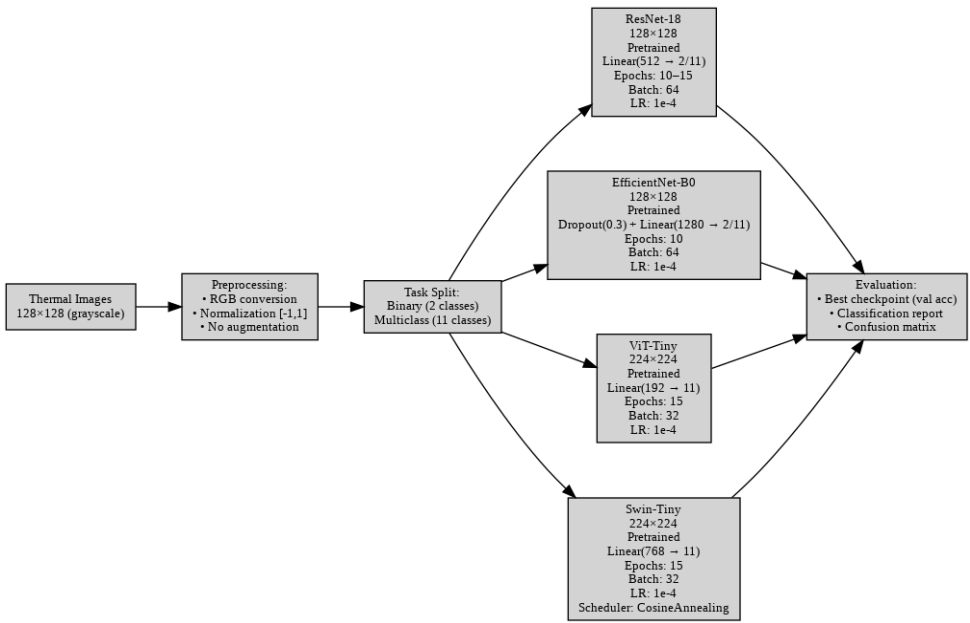

**Figure 4.** Experimental Setup Flowchart

ResNet-18 served as the base convolutional network with its final fully connected layer changed to produce 2 units for binary or 11 units for multiclass fault classification. The model used its pre-trained residual learning framework while adapting the classification head to accommodate thermal fault detection requirements.

EfficientNet-B0 involved compound scaling principles and necessitated adjustment of its classifier module. The initial classification layer was substituted by a sequential architecture with dropout regularization (p=0.3) prior to a linear layer with task-relevant output dimensions. This addition of dropout was specifically intended to counter overfitting issues in light of the comparatively limited thermal imagery dataset relative to natural image datasets.

ViT-Tiny used pure attention mechanisms via its transformer architecture. The initial classification token projection was exchanged for linear layers that corresponded to the target class dimensions without altering the model's self-attention mechanism and positional encoding schemes. This architecture facilitated the direct modeling of spatial relationships via attention instead of convolutional inductive biases.

Swin-Tiny employed hierarchical attention patterns combining transformer advantages with computational efficiency through shifted window mechanisms. Similar to ViT-Tiny, the classification head was adapted for thermal fault detection without disturbing the multi-scale feature processing capability of the model.

All models were pre-trained on ImageNet and transferred to thermal infrared imagery using careful architectural modifications. The thermal images, which were originally single-channel grayscale representing temperature distributions, were converted to three-channel via channel replication to make them compatible with RGB-trained backbones without losing thermal information content.

**Table 3.** Training Configuration for All Model–Task Combinations

| Model | Task | Epochs | Batch Size | Input Resolution | Learning Rate | Scheduler |
|---|---|---|---|---|---|---|
| **ResNet-18** | Binary | 10 | 64 | 128×128 | $1\times10^{-4}$ | None |
| **ResNet-18** | Multiclass | 15 | 64 | 128×128 | $1\times10^{-4}$ | None |
| **EfficientNet-B0** | Binary | 10 | 64 | 128×128 | $1\times10^{-4}$ | None |
| **EfficientNet-B0** | Multiclass | 10 | 64 | 128×128 | $1\times10^{-4}$ | None |
| **ViT-Tiny** | Binary | 15 | 32 | 224×224 | $1\times10^{-4}$ | None |
| **ViT-Tiny** | Multiclass | 15 | 32 | 224×224 | $1\times10^{-4}$ | None |
| **Swin-Tiny** | Binary | 15 | 32 | 224×224 | $1\times10^{-4}$ | CosineAnnealingWarmRestarts |
| **Swin-Tiny** | Multiclass | 15 | 32 | 224×224 | $1\times10^{-4}$ | CosineAnnealingWarmRestarts |

All models employed Adam optimizer with a learning rate of $1\times10^{-4}$ and cross-entropy loss for both classification. The training process differed across its architecture families based on their convergence behavior and need for adaptation. Convolutional models (ResNet-18 and EfficientNet-B0) were trained for 10-15 epochs based on their rapid feature learning capability and faster convergence while adapting from natural images to thermal domain. The compound scaling principles and residual learning architecture enable quick adaptation of pre-trained features with relatively few training iterations (Table 1).

Transformer models, specifically ViT-Tiny and Swin-Tiny, needed longer training durations of 15 epochs for each task conducted. This was because of intrinsic differences in how attention mechanisms encode spatial information differently than convolutional operations. This is because, in general, transformers need more iterations to align their learned attention patterns from natural images to the unique spatial and spectral features intrinsic in thermal fault signatures.

Batch size tuning was crafted to match computational requirements and optimization characteristics of each architecture family. Convolutional architectures utilized batch sizes of 64, leveraging their memory efficiency and stable gradient behavior. Transformer architectures adopted smaller batch sizes of 32 due to higher memory requirements.

Learning rate scheduling was used selectively based on architecture-specific requirements. Most models employed fixed learning rates for the sake of comparison consistency. Swin-Tiny, however, employed CosineAnnealingWarmRestarts scheduling with initial restart interval $T_0=5$ and multiplying factor T_mult=2, since this cyclic approach is found to be especially effective for hierarchical attention frameworks for domain adaptation.

Input resolution requirements varied between architecture families due to their design and pre-training regimes. Convolutional models natively processed images at their original resolution without explicit resizing, since the native spatial detail of thermal images was retained. Transformer-based models had to be resized to 224×224 pixels to align with their pre-training resolution and positional encoding mechanisms for the best performance of their attention mechanisms.

Thermal images underwent minimal preprocessing so as not to tamper with the integrity of temperature-based signatures. Normalization employed channel-wise mean and standard deviation of 0.5, scaling pixel intensities to the range [-1, 1]. Symmetric normalization over ImageNet statistics was preferred for its better alignment with thermal data characteristics and for avoiding the introduction of bias from natural image distributions.

Geometric and photometric augmentation strategies were purposely omitted from the preprocessing pipeline because of the sensitivity of thermal patterns to spatial and intensity changes. Traditional augmentation techniques may perturb the exact temperature relationships that define various fault types, thereby potentially degrading model performance or creating artifacts not typical of actual thermal signatures.

The dataset employed stratified random sampling for train-validation-test splits with proportions of 70-15-15 and with equal class distribution in subsets. Stratification in this way provided representative coverage of fault types in each partition without inducing class imbalance problems that would distort model assessment.

Model selection used validation-based early stopping, monitoring validation accuracy to select optimal model states. The best-performing weights found via validation performance were saved for final test set evaluation to guarantee reported metrics demonstrate maximum model capabilities rather than potentially degraded performance from overtraining.

An analysis employed classification reports and confusion matrices derived from the held-out test sets using the optimal model weights. The performance metrics incorporated precision, recall, F1-score, and overall accuracy computed per category, enabling a thorough understanding of the model's performance on various fault classifications and facilitating the identification of challenges applicable to thermal fault detection classification.

The experiment setup utilized PyTorch for model implementation and training with CUDA acceleration for GPU utilization efficiency. Transformer architectures were taken from the timm library to ensure availability of optimized implementations with pre-trained weights, while torchvision implementations were used for convolutional models. Data handling was implemented with NumPy for numerical operations and PIL for image processing, and evaluation metric computation was carried out with scikit-learn utilities and visualization with matplotlib and seaborn libraries.

## 3. Results

### 3.1 Binary Classification Performance

All four architectures demonstrated strong performance on the binary anomaly detection task, achieving accuracies between 93-94% on the test set. The results reveal remarkable consistency across different architectural paradigms, suggesting that the thermal fault detection problem is well-suited to deep learning approaches regardless of the underlying feature extraction mechanism (Table 4).

**Table 4.** Binary Classification Performance

| Model | No-Anomaly Precision | No-Anomaly Recall | No-Anomaly F1 | Faulty Precision | Faulty Recall | Faulty F1 | Accuracy | Macro Avg F1 |
|---|---|---|---|---|---|---|---|---|
| **ResNet-18** | 0.91 | 0.95 | 0.93 | 0.94 | 0.91 | 0.92 | 0.93 | 0.93 |
| **EfficientNet-B0** | 0.91 | 0.96 | 0.93 | 0.95 | 0.91 | 0.93 | 0.93 | 0.93 |
| **ViT** | 0.93 | 0.95 | 0.94 | 0.95 | 0.93 | 0.94 | 0.94 | 0.94 |
| **Swin** | 0.93 | 0.95 | 0.94 | 0.95 | 0.93 | 0.94 | 0.94 | 0.94 |

The transformer models, ViT-Tiny and Swin-Tiny showed a slightly improved performance level, with 94% accuracy compared to the 93% accuracy posted by the convolutional models, ResNet-18 and EfficientNet-B0. While the improvement is subtle, it highlights the effectiveness of attention mechanisms in capturing subtle thermal signatures that distinguish functional solar panels from faulty ones. An analytical comparison with a confusion matrix shows that all models show similar performance on both classes, with false positive rates (normal panels as faulty) at 4% to 5%, and false negative rates (faulty panels as normal) at 5% to 7%.

EfficientNet-B0 showed the best precision for faulty panels (0.95), reflecting a greater ability to limit false positives in the detection of panels that need maintenance. Conversely, this model achieved the highest recall rate for normal panels (0.96), reflecting excellent sensitivity for the identification of normally functioning panels. The transformer models presented more balanced precision-recall characteristics, with both metrics reaching 0.93-0.95 values for the various categories.

## 3.2 Multiclass Fault Classification Performance

The classification of faults into eleven different classes was even more challenging, as shown by overall accuracy rates ranging from 69% for the convolutional models to 73% for Swin-Tiny. Such variation in performance highlights the increased difficulty of distinguishing between eleven different fault classes compared to simple binary anomaly detection (Table 5).

**Table 5.** Multiclass Classification Performance

| Model | Accuracy | Macro Avg F1 | Weighted Avg F1 |
|---|---|---|---|
| **ResNet-18** | 0.69 | 0.63 | 0.69 |
| **EfficientNet-B0** | 0.69 | 0.62 | 0.68 |
| **ViT** | 0.72 | 0.68 | 0.72 |
| **Swin** | 0.73 | 0.70 | 0.73 |

Swin-Tiny is recognized as the best-performing model in multiclass classification, recording a 73% accuracy rate as well as a macro-averaged F1-score of 0.70. This outstanding performance is largely attributed to the hierarchical attention mechanism that efficiently handles multi-scale features, displaying significant strengths in distinguishing visually similar fault classes. ViT-Tiny also showed impressive performance, recording an accuracy of 72%, thus further emphasizing the overall effectiveness of transformer architectures in the area of thermal pattern recognition.

The convolutional architectures, ResNet-18 and EfficientNet-B0, achieved the same overall accuracy rate of 69%. This indicates that both methodologies, residual learning and compound scaling, face similar limitations to distinguishing nuanced thermal signatures. However, a deeper dive shows that the models excel in detecting some types of faults but fail to do so with others.

## 3.3 Per-Class Performance Analysis

A review of the performance on individual fault classes shows that there are significant differences in detection challenge between fault types. Diode faults (Class 4) showed consistently better classification accuracy for all models, reflected in precision and recall values greater than 0.90 for all architectures considered. This exceptional effectiveness can presumably be traced to the distinctive thermal signatures for diode faults, which produce characteristic hot spots that are easily discernible from other fault patterns.

**Table 6.** Multiclass Classification Performance (per Class)

| Model | Class | Precision | Recall | F1-Score |
|---|---|---|---|---|
| **ResNet-18** | Cell | 0.67 | 0.56 | 0.61 |
| | Hot-Spot | 0.57 | 0.54 | 0.56 |
| | Offline-Module | 0.73 | 0.81 | 0.77 |
| | Vegetation | 0.63 | 0.75 | 0.69 |
| | Diode | 0.94 | 0.92 | 0.93 |
| | Shadowing | 0.82 | 0.81 | 0.82 |
| | Cracking | 0.66 | 0.69 | 0.67 |
| | Diode-Multi | 0.80 | 0.62 | 0.70 |
| | Hot-Spot-Multi | 0.51 | 0.57 | 0.54 |
| | Cell-Multi | 0.50 | 0.50 | 0.50 |
| | Soiling | 0.25 | 0.16 | 0.20 |
| **EfficientNet-B0** | Cell | 0.68 | 0.65 | 0.66 |
| | Hot-Spot | 0.56 | 0.51 | 0.54 |
| | Offline-Module | 0.68 | 0.80 | 0.73 |
| | Vegetation | 0.67 | 0.73 | 0.70 |
| | Diode | 0.91 | 0.91 | 0.91 |
| | Shadowing | 0.76 | 0.76 | 0.76 |
| | Cracking | 0.66 | 0.73 | 0.69 |
| | Diode-Multi | 0.83 | 0.58 | 0.68 |
| | Hot-Spot-Multi | 0.47 | 0.51 | 0.49 |
| | Cell-Multi | 0.49 | 0.42 | 0.45 |
| | Soiling | 0.35 | 0.19 | 0.25 |
| **ViT** | Cell | 0.68 | 0.65 | 0.67 |
| | Hot-Spot | 0.64 | 0.68 | 0.66 |
| | Offline-Module | 0.77 | 0.85 | 0.81 |
| | Vegetation | 0.74 | 0.74 | 0.74 |
| | Diode | 0.97 | 0.95 | 0.96 |
| | Shadowing | 0.78 | 0.85 | 0.81 |
| | Cracking | 0.67 | 0.79 | 0.72 |
| | Diode-Multi | 0.91 | 0.81 | 0.86 |
| | Hot-Spot-Multi | 0.56 | 0.49 | 0.52 |
| | Cell-Multi | 0.49 | 0.42 | 0.45 |
| | Soiling | 0.33 | 0.32 | 0.33 |
| **Swin** | Cell | 0.70 | 0.72 | 0.71 |
| | Hot-Spot | 0.69 | 0.68 | 0.68 |
| | Offline-Module | 0.80 | 0.88 | 0.84 |
| | Vegetation | 0.74 | 0.67 | 0.70 |
| | Diode | 0.96 | 0.96 | 0.96 |
| | Shadowing | 0.85 | 0.82 | 0.83 |
| | Cracking | 0.70 | 0.68 | 0.69 |
| | Diode-Multi | 0.92 | 0.85 | 0.88 |
| | Hot-Spot-Multi | 0.56 | 0.62 | 0.59 |
| | Cell-Multi | 0.47 | 0.53 | 0.50 |
| | Soiling | 0.35 | 0.23 | 0.27 |

Fault classes are categorized into three levels of performance based on their F1-scores on various models. The high-performance fault classes with F1-scores of more than 0.80 are diode faults, whose scores ranged from 0.93 for ResNet-18 to 0.96 for ViT-Tiny; offline module faults, whose scores ranged from 0.73 for EfficientNet-B0 to 0.84 for Swin-Tiny; and shadowing faults, with scores ranging from 0.67 for ResNet-18 to 0.83 for Swin-Tiny. The medium-performance fault classes with F1-scores ranging from 0.60 to 0.80 are cell faults with relatively uniform performance levels ranging from 0.61 to 0.71 on the models; vegetation faults with scores ranging from 0.69 for ResNet-18 to 0.74 for ViT-Tiny; and cracking faults with scores ranging from 0.67 for ResNet-18 to 0.72 for ViT-Tiny. The most difficult-to-detect fault classes with F1-scores of less than 0.60 are soiling faults, which had the poorest performance of all models with scores ranging from 0.20 to 0.33; cell-multi faults, which were consistently difficult with scores ranging from 0.45 to 0.50; and hot-spot variants, that is, hot-spot and hot-spot-multi faults, whose performance varied with F1-scores ranging from 0.49 to 0.68.

These performance hierarchies reflect the operational criticality of different fault types in photovoltaic installations. Electrical faults achieving excellent detection rates (F1-scores >0.80) align with industrial data showing that electrical component failures cause disproportionately high energy losses despite lower occurrence frequency, with individual electrical faults contributing 0.3-1.5% daily production losses compared to module-level failures[2]. The superior detection of diode and offline-module faults therefore addresses the fault categories with greatest economic impact per incident.

**3.4 Error Analysis and Misclassification Patterns**

**Table 7.** Multiclass Confusion Matrices by Model

| Model | Multiclass Confusion Matrix |
|---|---|
| **ResNet-18** | 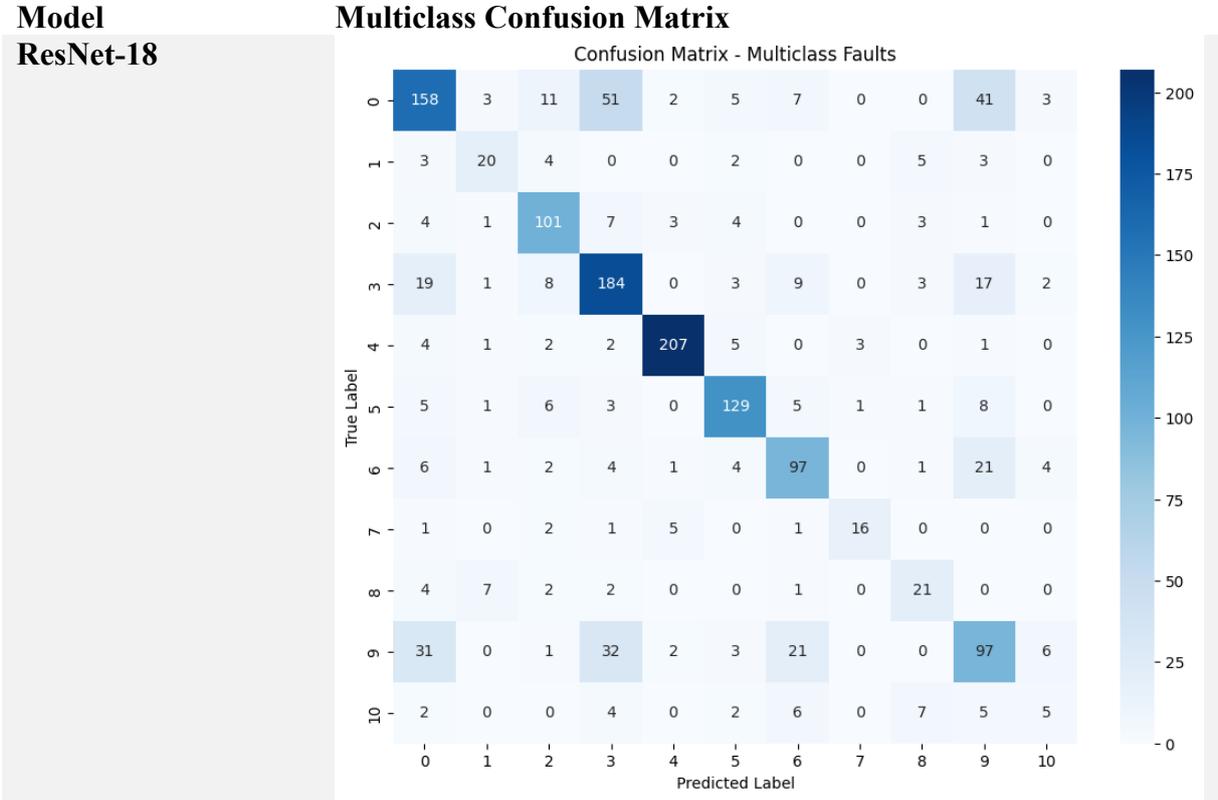 |

**EfficientNet-B0**

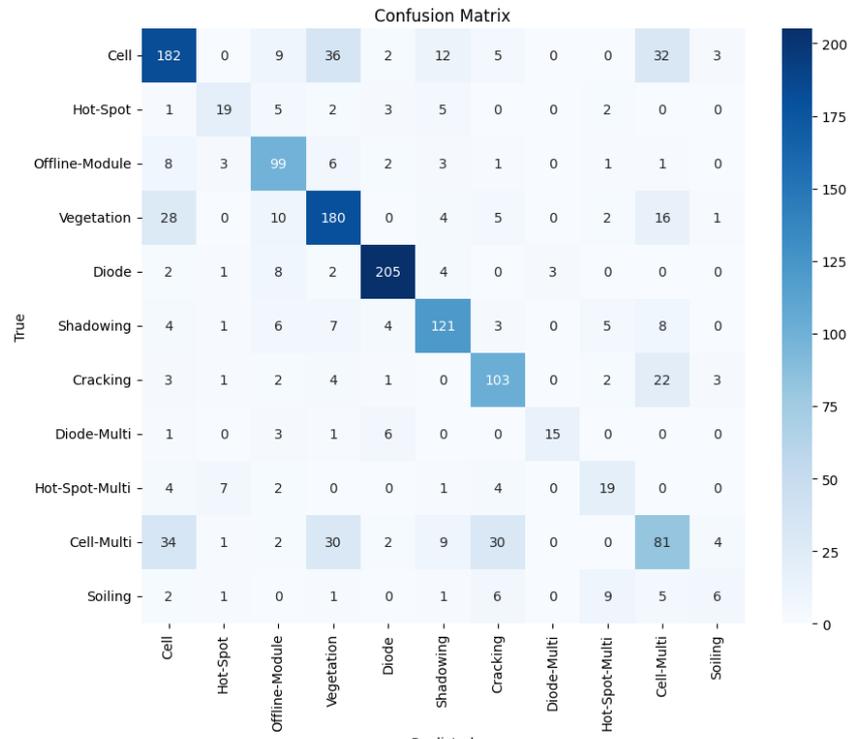

**ViT**

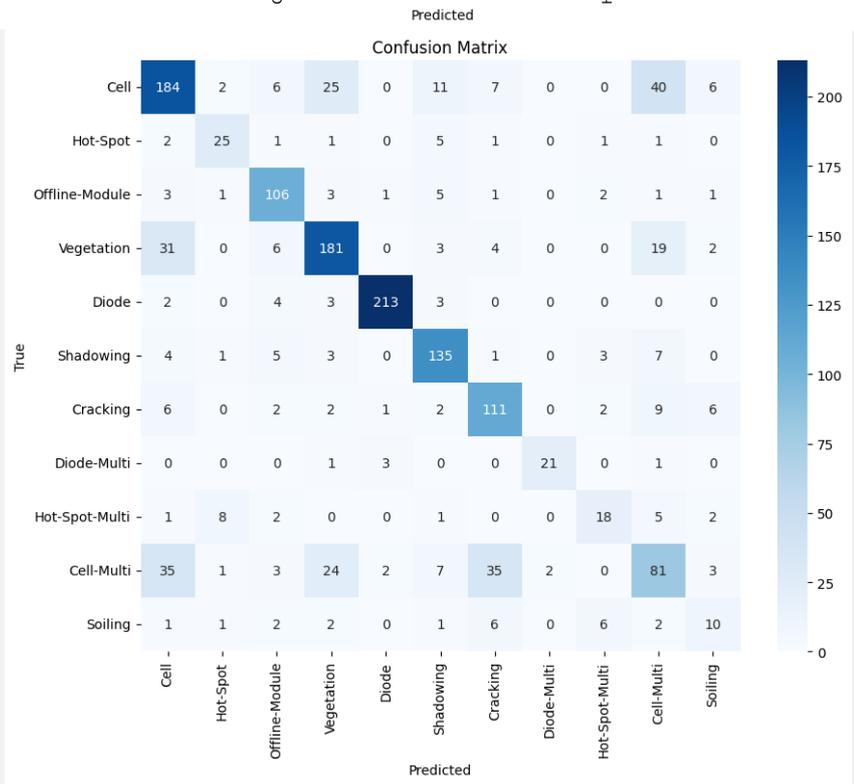

**Swin**

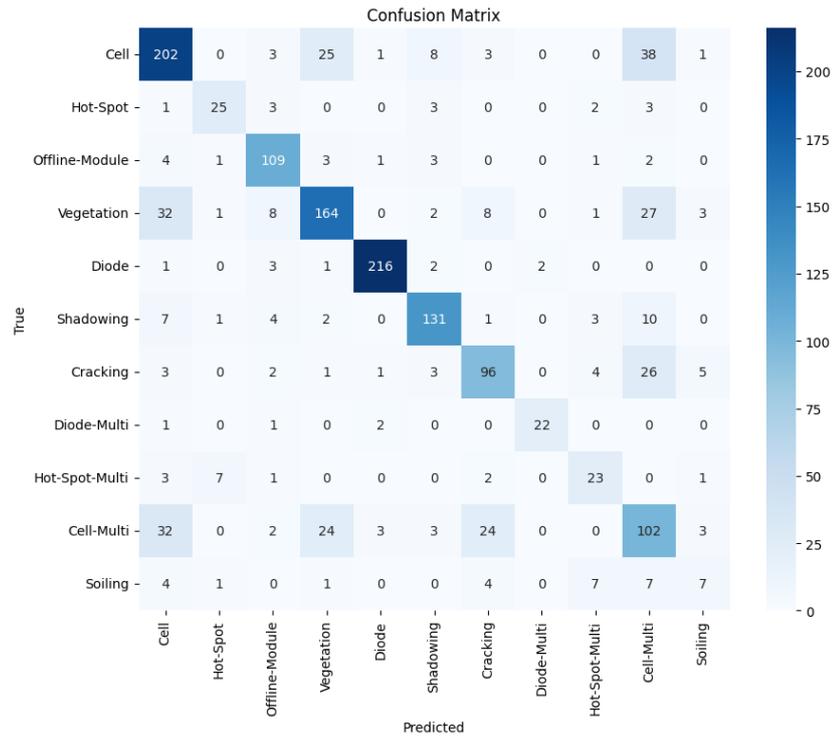

Analysis of confusion matrices reveals patterns of consistent misclassification that emphasize the inherent challenges of thermal fault detection. The common errors are caused by the misattribution of fault classes with similar thermal signatures, particularly the frequent misclassification of cell faults (Class 0) as cell-multi faults (Class 9) and the confusion of vegetation faults (Class 3) with cell-class-related faults. Cross-category confusion largely concerns a number of different patterns that identify intrinsic challenges in the interpretation of thermal signatures. Single-cell fault errors (Class 0) often show high confusion with vegetation faults (Class 3) and multi-cell faults (Class 9), suggesting difficulty in distinguishing isolated cell defects from environmental or multiple cell conditions. Soiling faults (Class 10) show the widest confusion pattern, being misclassified in many different categories, including cell faults, vegetation, and cracking, highlighting the inherently subtle nature of thermal changes caused by dust. Hot-spot faults (Class 1) and hot-spot-multi faults (Class 8) show moderate levels of confusion with each other and classes involving cells, revealing problems in measuring the magnitude and severity of hot spots. The observed error patterns suggest that future improvements might focus on developing specialized attention mechanisms or multi-scale feature extraction techniques tailored specifically to the inspection of thermal signatures. The ongoing difficulty of detecting soiling on different models, illustrated by its consistent misclassification with other fault categories, suggests an inherent limitation that might justify research into other sensing modalities, the advancement of thermal imaging resolution, or the integration with auxiliary detection methods, such as visual spectrum analysis or impedance testing.

### 3.5 XRAI Interpretability Analysis

### 3.5.1 XRAI Saliency Map Evaluation

To understand the decision-making process of the highest-performing Swin-Tiny model, XRAI (eXplainable AI for Region-based Analysis) generated saliency maps for correctly classified instances in each fault category. This analysis shows the particular image regions that the model considers most relevant to its classification decisions, thus providing valuable insights into the learned feature representations and their relation to the underlying principles of thermal physics.

**Table 8.** XRAI Saliency Analysis by Fault Class

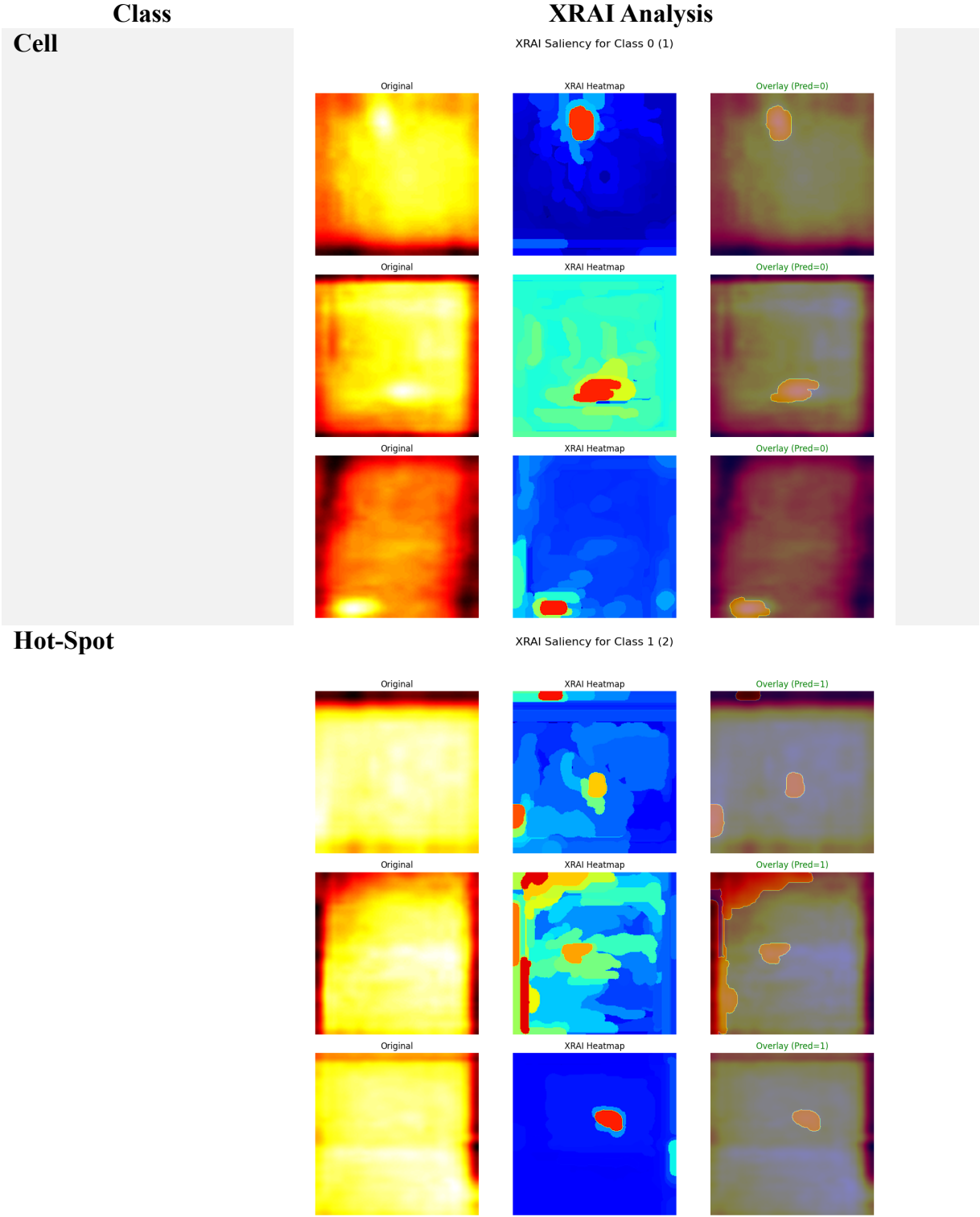

**Offline-Module**

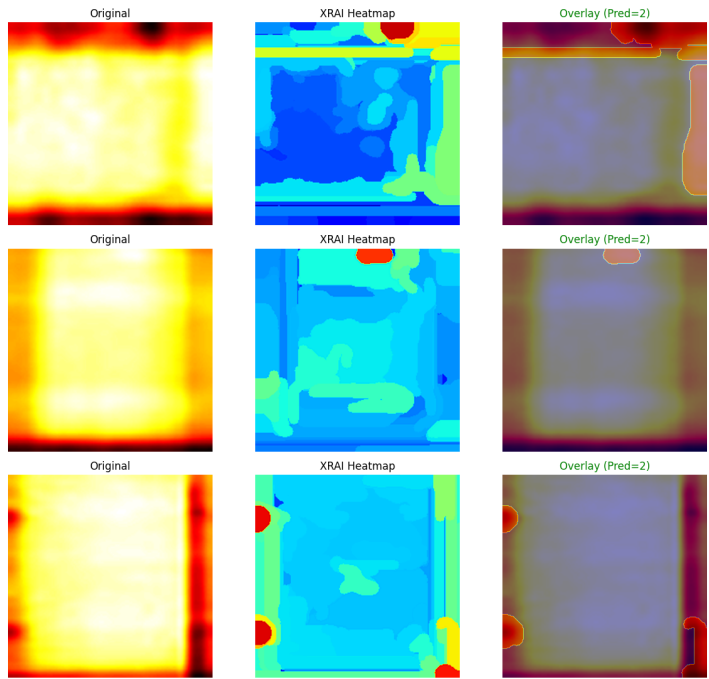

**Vegetation**

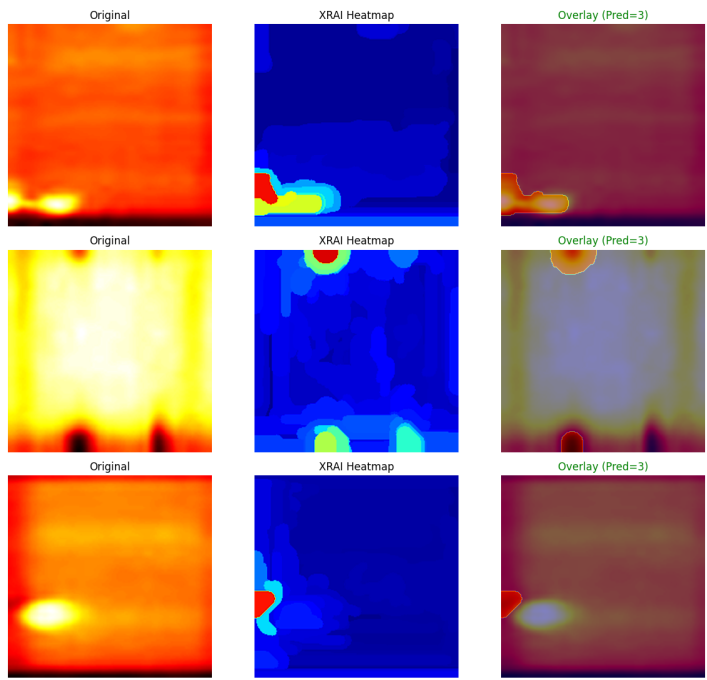

**Diode**

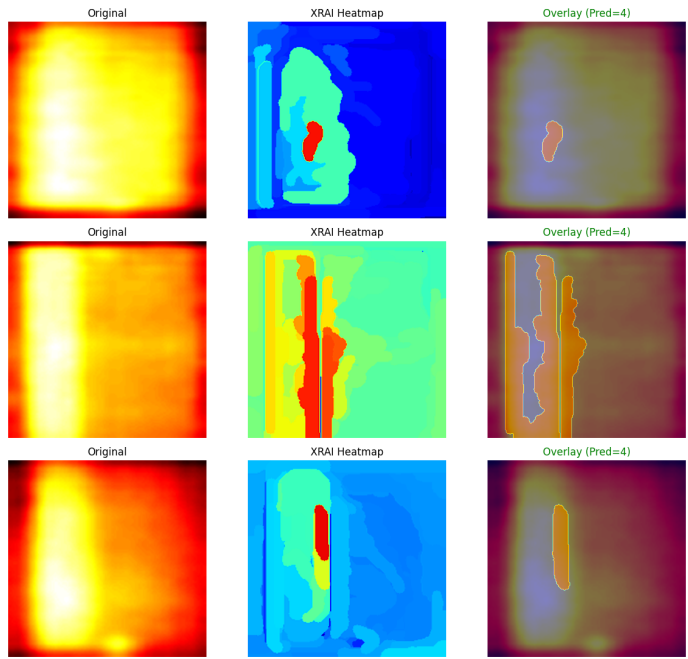

**Shadowing**

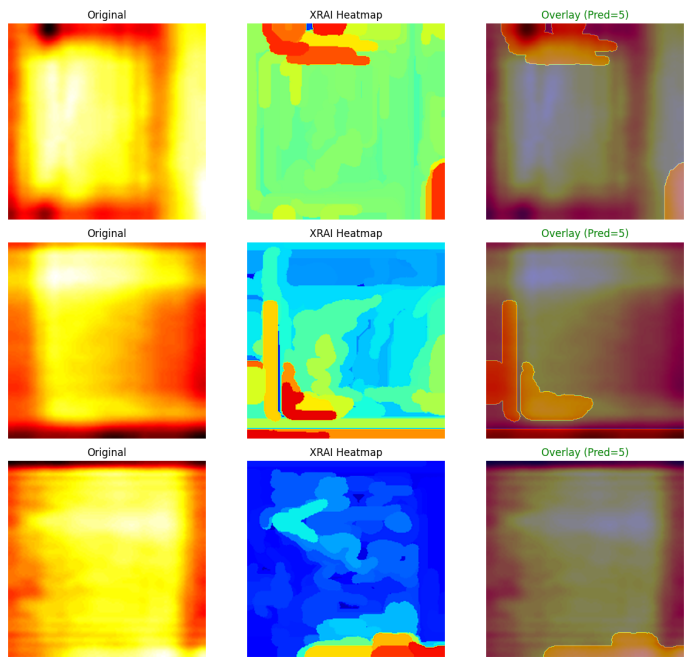

**Cracking**

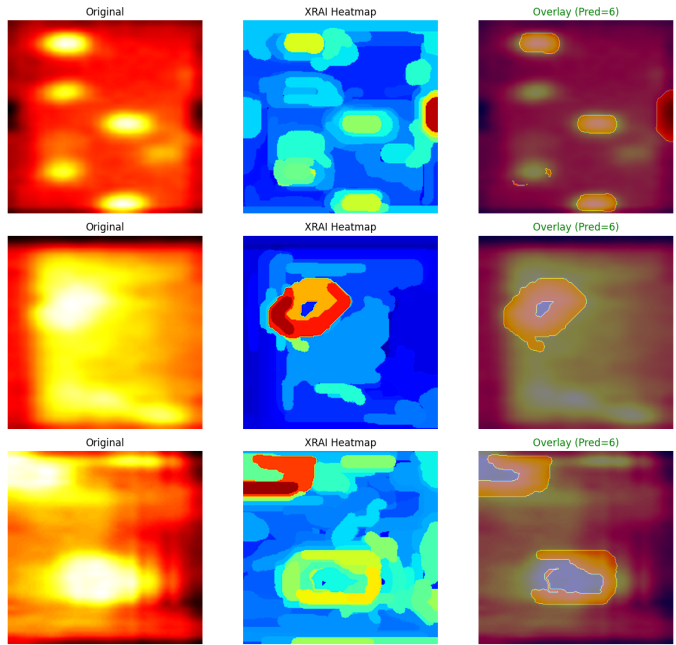

**Diode-Multi**

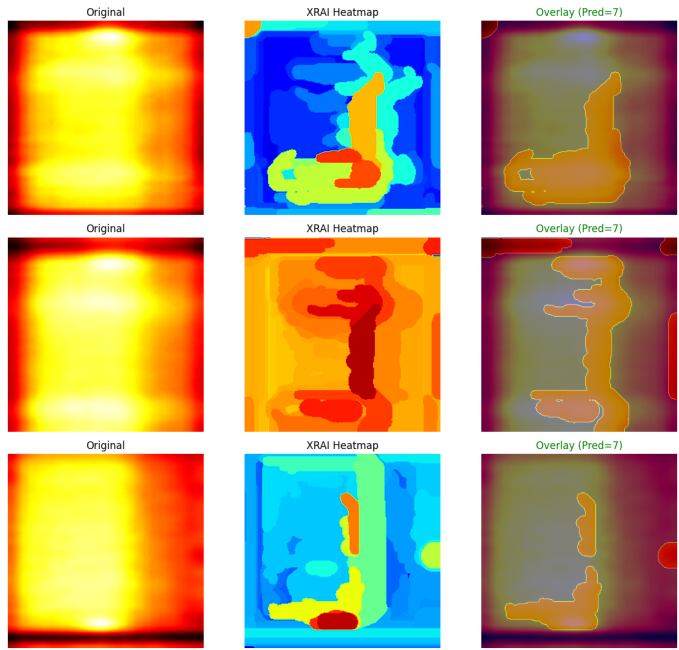

**Hot-Spot-Multi**

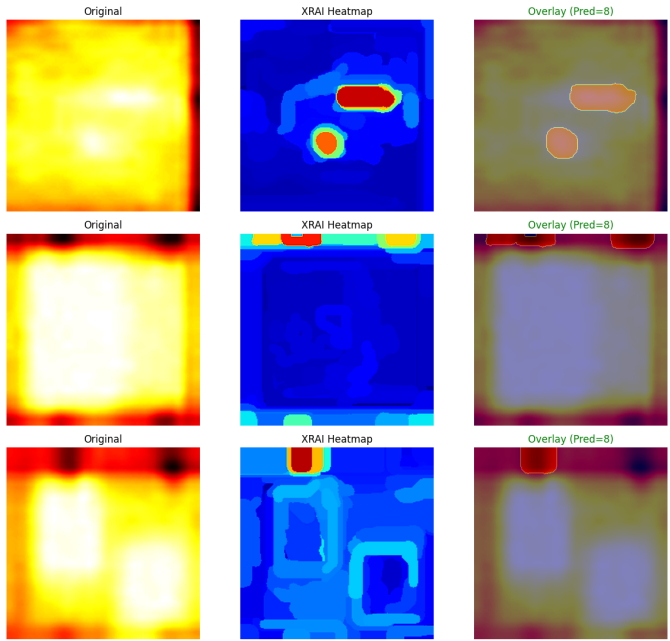

**Cell-Multi**

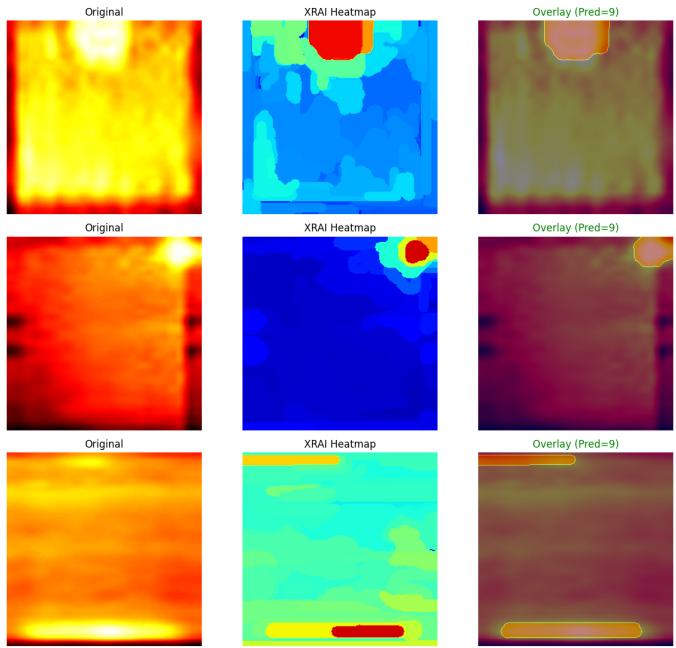

**Soiling**

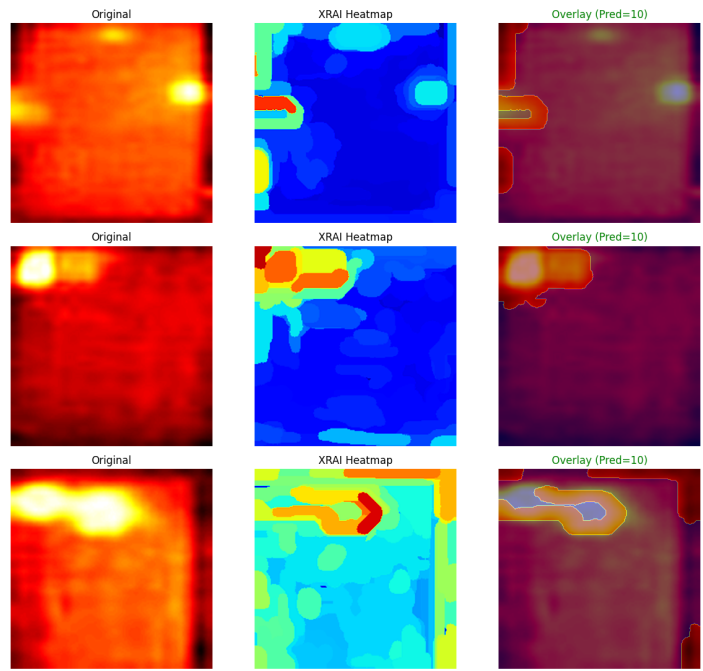

### 3.5.2 Physically Meaningful Attention Patterns

The XRAI test shows that the model attains a strong understanding of physical concepts related to different types of faults. For Cell faults (Class 0), the model always focuses its attention at precise hotspots in individual solar cells, correctly corresponding to the expected thermal patterns characteristic of electrical faults at the cellular level. The attention maps show precise localization to round or oval areas that correspond to individual faulty cells, thus showing that the model has indeed learned to distinguish between the characteristic thermal signatures that mark cell degradation, as opposed to relying on misleading visual cues (Table 8).

Diode faults (Class 4) present the most striking results in terms of interpretability, with XRAI maps showing distinct vertical or linear hot areas in direct association with bypass diode locations and current paths. The model's attention mechanism correctly identifies the characteristic linear thermal patterns produced by diode faults, which manifest as long hot areas along the edges of the cells. This demonstrates a sophisticated understanding of both the electrical design of solar panels and the laws of thermal physics.

Class 1 hot-spot faults show strong localization features, where the model concentrates on small but important thermal anomalies corresponding to localized overheating due to the effects of electrical resistance or shading. The XRAI maps show salient, round areas of interest that relate to the expected thermal patterns related to the onset of hot spots.

Class 6 cracking faults display truly interesting features, since attention patterns follow linear patterns that relate to the physical paths of crack growth. The model has developed the capacity to identify not just the momentary thermal signatures of cracks, but also the related distributions of thermal stress, which demonstrate a sophisticated understanding of the mechanical failure modes of photovoltaic systems.

### 3.5.3 Reasonable but Complex Attention Patterns

Class 2 Offline-Module faults have attention patterns that are more complex but still within the bounds of physical reality. The model highlights peripheral regions and intersections, consistent with the expected thermal signatures due to electrical interruptions at the module level. However, the attention patterns are more scattered compared to what is present in particular fault classes, highlighting the systemic nature of offline module failures potentially caused by multiple thermal paths.

Shadowing faults (Class 5) display varied attentional patterns consistent with shadow edges and regions under partial shading. The model has developed the ability to identify not just the cooler regions that are shaded but also the compensatory warming in adjacent unshaded regions, hence demonstrating an awareness of the complex thermal interactions triggered by partial shading conditions.

Class 3 vegetation faults reflect attention patterns that prioritize thermal boundaries distinguishing areas with obstructed and unobstructed panels, correctly identifying the thermal gradients that occur where organic matter blocks solar irradiance.

### 3.5.4 Challenging Interpretability Cases

Interpretation of multiple fault classes, namely Classes 7, 8, and 9, is a big challenge. XRAI maps for Diode-Multi (Class 7), Hot-Spot-Multi (Class 8), and Cell-Multi (Class 9) reflect attention distributions over different parts of the panel, which fits logically given that these classes indicate the presence of concomitant failures in multiple locations. However, attention patterns in these classes are less localized compared to what happens in single-fault classes, making model validation as to whether the model is correctly picking up each unique fault component or simply picking up statistical relationships between multiple thermal anomalies an issue.

Soiling faults (Class 10) produce especially concerning results in terms of their interpretability. The attention patterns learned from XRAI for soiling show significant variability, often focusing on the periphery or single spots instead of the expected uniform reduction in thermal activity across surfaces impacted by dust. This observation suggests that the model may be learning to detect soiling by way of secondary thermal responses or visual anomalies, instead of the direct physical process of reduced solar absorption by accumulated dust. This realization is reflected in the poor classification performance documented for soiling faults and represents an important limitation in the detectability of weak diffuse thermal changes under existing image quality.

## 4. Discussion

### 4.1 Addressing Industrial Deployment Barriers in Solar Energy

The integration of artificial intelligence into solar energy infrastructure has been hindered by a fundamental deployment barrier, which is the inability to validate that AI systems make decisions based on physically meaningful thermal signatures rather than spurious correlations. This study demonstrates that high-performing models can learn thermally consistent features, directly addressing stakeholder confidence issues that have limited AI adoption in energy infrastructure.

Our XRAI analysis reveals that transformer attention mechanisms align with established thermal physics principles: localized hotspots for cell defects, linear thermal paths for diode failures, and boundary effects for vegetation shading. This represents a critical advance over previous work using this dataset. While Le et al. (2023) achieved superior multiclass accuracy (85.4%) using CNN-PVE and implemented Grad-CAM++ visualization, their analysis focused on general feature localization rather than systematic validation against thermal physics principles[18]. Similarly, Daher et al. (2024) combined multiple inspection modalities and V et al. (2023) evaluated individual CNN architectures, but none validated model decisions against domain physics[19,20].

This physics-grounded validation addresses deployment barriers in an economically critical context, where equipment failures contribute measurable energy losses reaching 0.96% of net energy yield in operational installations, with electrical faults accounting for 68% of total failure energy losses despite representing <7% of total failures[2]. The demonstrated alignment between model attention and thermal physics principles provides stakeholders with confidence in AI-driven maintenance prioritization for these high-impact failure modes.

### 4.2 Architectural Trade-offs and Energy Infrastructure Deployment Strategies

The systematic architectural comparison reveals deployment trade-offs critical for solar energy operations. Transformer models (ViT-Tiny: 72%, Swin-Tiny: 73%) outperformed CNNs (ResNet-18, EfficientNet-B0: 69%) in multiclass fault classification, though none matched the 85.4% achieved by Le et al.'s specialized CNN-PVE architecture. However, our transformer approaches demonstrate good binary anomaly detection performance (94% vs 93% for CNNs) with the crucial addition of physics-validated interpretability absent in previous work.

Our comprehensive evaluation addresses gaps in thermal infrastructure monitoring identified in prior studies. While V et al. (2023) and other researchers using this dataset focused on individual CNN architectures without systematic comparison, our work reveals that architectural choice represents a fundamental trade-off between computational efficiency and detection accuracy that should inform deployment decisions in solar energy operations[20].

These performance differences translate to distinct deployment scenarios in solar energy infrastructure. CNNs offer computational efficiency essential for distributed monitoring systems, drone-based inspections, embedded sensors in remote installations, and real-time edge processing where power and bandwidth constraints dominate. Their 93% binary accuracy suffices for first-stage anomaly screening while maintaining operational feasibility in resource-limited environments.

Transformers prove optimal for centralized diagnostic systems where comprehensive fault assessment justifies computational overhead. In utility-scale operations managing hundreds of thousands of panels, the enhanced multiclass discrimination capability and physics-aligned attention patterns support precision-critical maintenance decisions. The hierarchical attention in Swin-Tiny effectively processes both localized thermal anomalies and global panel-level patterns essential for accurate fault assessment in large installations.

### 4.3 Operational Performance Analysis and Industry Implications

Performance disparities across fault categories reveal fundamental limitations with direct operational implications. Electrical faults with distinct thermal signatures enable reliable automated detection that prevents high-consequence failures, while environmental faults present persistent challenges constraining automated monitoring capabilities. Analysis of operational installations confirms that electrical failures cause substantially higher per-incident energy losses (0.3-1.5% daily production) compared to environmental degradation, validating the economic priority of achieving excellent electrical fault detection over comprehensive environmental fault coverage[2]. However, soiling's potential to reduce panel output by 20-40% annually in desert installations means that thermal-only approaches face inherent detection limits for economically significant environmental degradation.

Environmental faults present persistent challenges that constrain automated monitoring capabilities. Soiling detection achieved only 0.20-0.33 F1-scores across all models, consistent with limitations reported by Le et al. (2023), indicating fundamental constraints in thermal-only approaches for dust accumulation detection[18]. This limitation has significant operational implications: soiling can reduce panel output by 20-40% annually in desert installations, yet current thermal imaging approaches cannot reliably detect this economically critical fault type.

Confusion matrix analysis reveals systematic misclassification patterns affecting maintenance operations. Cell faults frequently confused with multi-cell conditions and vegetation faults misattributed to cell-related problems indicate that thermal imaging discrimination capabilities constrain maintenance precision. The 40×24 pixel resolution with substantial pixelation fundamentally limits fine-grained feature extraction necessary for distinguishing subtle environmental degradation, precisely the fault types where early detection provides maximum economic benefit through preventive intervention.

These findings suggest that thermal-only monitoring approaches have reached physical detection limits for environmental fault types. Cost-effective solar energy operations may require hybrid sensing strategies combining thermal imaging for electrical fault detection with complementary modalities, such as optical inspection for soiling assessment, electrical impedance measurement for cell degradation monitoring to achieve comprehensive coverage across the fault spectrum affecting photovoltaic installations.

### 4.4 Framework Transferability to Critical Energy Infrastructure

The physics-validated interpretability methodology addresses deployment barriers extending across energy infrastructure domains where model explainability determines stakeholder acceptance. Power grid thermal monitoring, wind turbine condition assessment, and energy storage system diagnostics share common requirements: AI decisions must align with established engineering principles to support operational integration.

The demonstrated approach provides a transferable template for validating AI decisions against domain-specific physics principles. Power grid applications could validate thermal anomaly detection against electrical load theory, wind systems could verify bearing fault detection against mechanical failure physics, and battery systems could confirm degradation assessment against electrochemical principles. This domain-specific validation approach addresses regulatory and operational barriers that have limited AI adoption in safety-critical energy systems despite demonstrated technical capabilities.

The realistic class imbalance in our evaluation, soiling representing only 2.0% of samples, provides operationally relevant assessment reflecting actual fault distributions in deployed installations. This realistic evaluation demonstrates that transformer architectures maintain performance advantages under operational conditions, supporting deployment viability in real-world energy infrastructure scenarios.

**4.5 Research Directions for Energy Infrastructure AI**

Future research priorities should address the identified thermal imaging limitations through multi-modal sensing integration. Combining thermal data with high-resolution optical imaging, electrical parameter monitoring, and environmental sensor data could overcome single-modality constraints while maintaining the physics-validated interpretability framework demonstrated in this work.

Physics-informed neural architectures represent a promising direction for incorporating domain knowledge directly into model structures. Such approaches could embed thermal transfer equations, electrical circuit principles, and mechanical failure modes into network architectures, potentially improving both performance and interpretability while reducing training data requirements, a significant advantage for energy applications where fault examples are naturally rare.

The regulatory implications require investigation regarding validation protocols for AI systems in critical energy infrastructure. Establishing standardized approaches for physics-grounded interpretability validation could facilitate broader AI adoption while ensuring appropriate safety and reliability standards essential for energy system integration. Industry collaboration between AI researchers and energy system engineers is crucial for developing validation frameworks that satisfy both technical performance and operational acceptance requirements.

# 5. Conclusion

This research established a comprehensive benchmark for deep learning-based thermal solar panel fault detection, with Swin Transformer achieving a performance of 94% binary and 73% multiclass accuracy. The novel XRAI interpretability analysis validated that model decisions align with thermal physics principles, addressing critical deployment concerns for industrial AI systems. Significant performance variations across fault categories were identified, with electrical faults achieving excellent detection while environmental factors like soiling present ongoing challenges. The approach addresses critical deployment barriers where equipment failures contribute up to 0.96% energy losses in operational installations, with validated model decisions focusing on fault categories responsible for the majority of failure-related economic impact. The demonstrated competencies enable the swift integration into maintenance procedures through automated drone inspection and focused maintenance methods. Image quality-related constraints were identified as the primary bottleneck to further improvement, suggesting that development in thermal imaging quality could dramatically expand diagnostic capabilities. The physics-validated approach provides a pathway for regulatory acceptance and industrial deployment where model interpretability is essential for stakeholder confidence. This work lays the groundwork for automated maintenance systems in solar power plants that ensure the expansion of renewable sources of energy, with performance metrics and interpretative methods translatable across other areas for thermal diagnostics.

## References

1. Spajić, M., Talajić, M. & Mršić, L. Using CNNs for Photovoltaic Panel Defect Detection via Infrared Thermography to Support Industry 4.0. *Business Systems Research Journal* **15**, 45–66 (2024).

2. Lillo-Bravo, I., González-Martínez, P., Larrañeta, M. & Guasumba-Codena, J. Impact of Energy Losses Due to Failures on Photovoltaic Plant Energy Balance. *Energies* **11**, 363 (2018).

3. Masita, K., Hasan, A., Shongwe, T. & Hilal, H. A. Deep learning in defects detection of PV modules: A review. *Solar Energy Advances* **5**, 100090 (2025).

4. Boubaker, S., Kamel, S., Ghazouani, N. & Mellit, A. Assessment of Machine and Deep Learning Approaches for Fault Diagnosis in Photovoltaic Systems Using Infrared Thermography. *Remote Sensing* **15**, 1686 (2023).

5. Bu, C., Liu, T., Wang, T., Zhang, H. & Sfarra, S. A CNN-Architecture-Based Photovoltaic Cell Fault Classification Method Using Thermographic Images. *Energies* **16**, 3749 (2023).

6. Dhimish, M. HOTSPOT-YOLO: A Lightweight Deep Learning Attention-Driven Model for Detecting Thermal Anomalies in Drone-Based Solar Photovoltaic Inspections. Preprint at https://doi.org/10.48550/ARXIV.2508.18912 (2025).

7. Bommes, L. *et al.* Anomaly Detection in IR Images of PV Modules using Supervised Contrastive Learning. Preprint at https://doi.org/10.48550/ARXIV.2112.02922 (2021).

8. Vlaminck, M., Heidbuchel, R., Philips, W. & Luong, H. Region-Based CNN for Anomaly Detection in PV Power Plants Using Aerial Imagery. *Sensors* **22**, 1244 (2022).

9. Li, B. *et al.* Trustworthy AI: From Principles to Practices. Preprint at https://doi.org/10.48550/ARXIV.2110.01167 (2021).

10. Wang, X. *et al.* A Survey on Trustworthy Edge Intelligence: From Security and Reliability To Transparency and Sustainability. Preprint at https://doi.org/10.48550/ARXIV.2310.17944 (2023).

11. Machlev, R. *et al.* Explainable Artificial Intelligence (XAI) techniques for energy and power systems: Review, challenges and opportunities. *Energy and AI* **9**, 100169 (2022).

12. Pelekis, S. *et al.* Trustworthy artificial intelligence in the energy sector: Landscape analysis and evaluation framework. Preprint at https://doi.org/10.48550/ARXIV.2412.07782 (2024).

13. Ledmaoui, Y., El Maghraoui, A., El Aroussi, M. & Saadane, R. Enhanced Fault Detection in Photovoltaic Panels Using CNN-Based Classification with PyQt5 Implementation. *Sensors* **24**, 7407 (2024).

14. Chung, N. C. *et al.* False Sense of Security in Explainable Artificial Intelligence (XAI). Preprint at https://doi.org/10.48550/ARXIV.2405.03820 (2024).

15. Ali, S. *et al.* Explainable Artificial Intelligence (XAI): What we know and what is left to attain Trustworthy Artificial Intelligence. *Information Fusion* **99**, 101805 (2023).

16. Riurean, S., Fîță, N.-D., Păsculescu, D. & Slușariuc, R. Securing Photovoltaic Systems as Critical Infrastructure: A Multi-Layered Assessment of Risk, Safety, and Cybersecurity. *Sustainability* **17**, 4397 (2025).

17. Millendorf, M., Obropta, E. & Vadhavkar, N. Infrared Solar Module Dataset for Anomaly Detection. in (2020).

18. Le, M., Le, D. & Ha Thi Vu, H. Thermal inspection of photovoltaic modules with deep convolutional neural networks on edge devices in AUV. *Measurement* **218**, 113135 (2023).

19. Daher, D. H. *et al.* Photovoltaic failure diagnosis using imaging techniques and electrical characterization. *EPJ Photovolt.* **15**, 25 (2024).

20. V, G. R. N., G, S. N., A, R. S., G, V. V. R. & Ch, Y. R. Anomaly Detection in Solar Modules with Infrared Imagery. *E3S Web Conf.* **391**, 01069 (2023).